\documentclass[prx,aps,twocolumn,amsmath,amssymb,longbibliography]{revtex4-1}

\pdfoutput=1
\usepackage{hyperref}
\usepackage{graphicx}
\usepackage{bm}
\usepackage[usenames]{color}

\def\cal#1{\mathcal{#1}}
\def\eqq#1{Eq.~(\ref{#1})}

\def\eq#1{(\ref{#1})}
\def\f#1{Fig.~\ref{#1}}

\def\s#1{Section~\ref{#1}}
\def\c#1{~\cite{#1}}

\def\cc#1{Ref.~\cite{#1}}

\def\av#1{\langle #1 \rangle}

\def\x{\bm x}
\def\e{\bm \epsilon}
\def\en{{\bm \epsilon} \cdot \nabla}

\def\p{P(\x,t)}
\def\pm{P(\x-\e,t)}
\def\w{W_{\e}(\x)}
\def\wm{W_{\e}(\x-\e)}
\def\wp{W(\x \to \x')}

\def\ee{{\rm e}}
\def\d{{\rm d}}
\def\u{U(\x)}
\def\uu{U}
\def\up{U(\x+\e)}
\def\beq{\begin{equation}}
\def\eeq{\end{equation}}
\def\bea{\begin{eqnarray}}
\def\eea{\end{eqnarray}}

\begin{document}

\title{Correspondence between neuroevolution and gradient descent}

\author{Stephen Whitelam$^1$}
\email{swhitelam@lbl.gov}
\author{Viktor Selin$^{2}$}
\author{Sang-Won Park$^1$}
\author{Isaac Tamblyn$^{2,3,4}$}
\email{ isaac.tamblyn@nrc.ca}
\affiliation{$^1$Molecular Foundry, Lawrence Berkeley National Laboratory, 1 Cyclotron Road, Berkeley, CA 94720, USA\\
$^2$Department of Physics, University of Ottawa, ON, K1N 6N5, Canada \\   
$^2$National Research Council of Canada, Ottawa, ON K1N 5A2, Canada\\
$^3$Vector Institute for Artificial Intelligence, Toronto, ON M5G 1M1, Canada}

\begin{abstract}
We show analytically that training a neural network by conditioned stochastic mutation or ``neuroevolution'' of its weights is equivalent, in the limit of small mutations, to gradient descent on the loss function in the presence of Gaussian white noise. Averaged over independent realizations of the learning process, neuroevolution is equivalent to gradient descent on the loss function. We use numerical simulation to show that this correspondence can be observed for finite mutations,for shallow and deep neural networks. Our results provide a connection between two families of neural-network training methods that are usually considered to be fundamentally different.
\end{abstract}
\maketitle

\section{Introduction}

In broad terms there are two types of method used to train neural networks, divided according to whether or not they explicitly evaluate gradients of the loss function. Gradient-based methods include the backpropagation algorithm\c{linnainmaa1976taylor,werbos1982applications,rumelhart1995backpropagation,rumelhart1986learning,hecht1992theory,lecun1989backpropagation,chauvin1995backpropagation,schmidhuber2015deep}. The non-gradient-based methods (sometimes called ``black box'' methods) include stochastic processes in which changes to a neural network are proposed and accepted with certain probabilities, and encompass Monte Carlo\c{metropolis1953equation,hastings1970monte} and genetic algorithms\c{GA,GA2,montana1989training}. Both gradient-based and non-gradient-based methods have been used to train neural networks for a variety of applications, and, where comparison exists, perform similarly well\c{DQN,morse2016simple,salimans2017evolution,zhang2017relationship}. For instance, recent numerical work shows that stochastic evolutionary strategies applied to neural networks are competitive with explicit gradient-based methods on hard reinforcement-learning problems\c{salimans2017evolution}.

Gradient-based- and non-gradient-based strategies are different in implementation and are sometimes thought of as entirely different approaches\c{sutton2018reinforcement}. Here we show that the two sets of methods have a fundamental connection. We demonstrate analytically an equivalence between the dynamics of neural-network training under conditioned stochastic mutations, and under gradient descent. This connection follows from one identified in the 1990s between the overdamped Langevin dynamics and Metropolis Monte Carlo dynamics of a particle in an external potential\c{kikuchi1991metropolis,kikuchi1992metropolis}. In the limit of small Monte Carlo trial moves, those things are equivalent. Similarly, we show here that a single copy of a neural network (a single individual) exposed to parameter mutations that are accepted probabilistically is equivalent, in the limit of small mutation size, to gradient descent on the loss function in the presence of Gaussian white noise. The details of the resulting dynamics depend on the details of the acceptance criterion, and encompass both standard- and clipped gradient descent. Such a mutation scheme corresponds to the simple limit of the set of processes called ``neuroevolution''\c{montana1989training,floreano2008neuroevolution,Guber,salimans2017evolution,whitelam2020learning}. This connection demonstrates explicitly that optimization without access to gradients can, nonetheless, enact noisy gradient descent on the loss function.

 In simple gradient descent, {equivalent to noise-free overdamped Langevin dynamics}, the parameters (weights and biases) $\x$ of a {neural} network evolve with training time according to the prescription $\dot{\x} =- \alpha \nabla \u$. Here $\alpha$ is a learning rate, and $\nabla \u$ is the gradient of a loss function $\u$ with respect to the network parameters. Now consider a simple neuroevolution scheme in which we propose a mutation $\x \to \x+ \e$ of all neural-network parameters, where $\e$ is a set of independent Gaussian random numbers of zero mean and variance $\sigma^2$. Let us accept the proposal with the Metropolis probability $\min\left(1,\ee^{-\beta \Delta U}\right)$. Here $\beta$ is a reciprocal temperature, and $\Delta U$ is the change of the loss function under the proposal. This is a {Metropolis Monte Carlo algorithm, a Markovian dynamics that constitutes a form of importance sampling}, and a common choice in the physics literature\c{metropolis1953equation,hastings1970monte,frenkel2001understanding}. In physical systems, $\beta$ {is inversely proportional to the physical temperature}, and we consider finite values of $\beta$ in order to make contact with that literature. However, in the context of training a neural network it is interesting to consider the zero-temperature limit $\beta = \infty$, where mutations are accepted only if the loss does not increase. That regime is not normally considered in particle-based simulations.

Our main results can be summarized as follows. When $\beta \neq \infty$ the weights of the network evolve, to leading order in $\sigma$, as $\dot{\x} = -(\beta \sigma^2/2) \nabla\u$ plus Gaussian white noise. Averaged over independent realizations of the learning process, {this form of neuroevolution is therefore equivalent to simple gradient descent}, with learning rate $\beta \sigma^2/2$. In the limit $\beta=\infty$, where mutations are accepted only if the loss function does not increase, weights under neuroevolution evolve instead as $\dot{\x}= - (\sigma/\sqrt{2 \pi})|\nabla \u|^{-1} \nabla \u$ plus Gaussian white noise, which corresponds to clipped gradient descent on $\u$\c{pascanu2013difficulty}. Note that conditioning the acceptance of neural-network parameter mutations on the change of the loss function for a single copy of that network is sufficient to enact gradient descent: a population of individuals is not required.

{In this paper we use the term ``neuroevolution'' to refer to a sequence of mutation steps applied to the parameters of a single copy of a neural network and accepted probabilistically. In general, neuroevolutionary algorithms encompass a broader variety of processes, including mutations of populations of communicating neural networks\c{salimans2017evolution} and mutations of network topologies\c{stanley2002evolving,floreano2008neuroevolution,stanley2019designing}. Similarly, the set of procedures for particles that can be described as ``Monte Carlo algorithms'' is large, and ranges from local moves of single particles -- roughly equivalent to the procedure used here -- to nonlocal moves and moves of collections of particles\c{swendsen1987nonuniversal,wolff1989collective,frenkel2001understanding,liu2004rejection,whitelam2011approximating}.) The dynamics of those collective-move Monte Carlo algorithms and of the more complicated neuroevolutionary schemes\c{stanley2002evolving,floreano2008neuroevolution,stanley2019designing} do not correspond to simple gradient descent. Here, we demonstrate a correspondence between one member of this set of algorithms and gradient descent, the implication being that, given any potentially complicated set of neuroevolutionary methods, it is enough to add simple a mutation-acceptance protocol in order to ensure that gradient descent is also approximated. The neuroevolution-gradient descent correspondence is similar to the proofs that neural networks with enough hidden nodes can represent any smooth function\c{cybenko1989approximation}: it does not necessarily suggest how to solve a given problem, but provides understanding of the limits and capacity of the tool and its relation to other methods of learning.}

{Our work provides a rigorous connection between gradient descent and what is arguably the simplest form of neuroevolution. It complements studies that demonstrate a numerical similarity between gradient-based methods and population-based evolutionary methods\c{salimans2017evolution,zhang2017relationship}, and studies that show analytically that the gradients approximated by those methods are, under certain conditions, equivalent to the finite-difference gradient\c{raisbeck2019evolution,staines2012variational, maheswaranathan2019guided}.}

{The paper is structured as follows.} We summarize the neuroevolution-gradient descent correspondence in \s{summary}, and derive it in \s{derivation}. Our derivation uses ideas developed in \cc{kikuchi1992metropolis} {to treat physical particles, and applies them to neural networks: we consider a different set-up~\footnote{In effect we work with a single particle in a high-dimensional space, rather than with many particles in three-dimensional space.} and proposal rates, and we consider the limit $\beta = \infty$ that is rarely considered in the physics literature but is natural in the context of a neural network. We can} associate the state $\x$ of the neural network with the position of a particle in a high-dimensional space, and the loss function $\u$ with an external potential. The result is a rewriting of the correspondence between Langevin dynamics and Monte Carlo dynamics as a correspondence between the simplest forms of gradient descent and neuroevolution.  Just as the Langevin-Monte Carlo correspondence provides a basis for understanding why Monte Carlo simulations of particles can approximate real dynamics\c{whitelam2007avoiding,wilber2007reversible,berthier2007revisiting,sanz2010dynamic,whitelam2011approximating,liu2013coarse,rovigatti2018simulate}, so the neuroevolution-gradient descent correspondence shows how we can effectively perform gradient descent on the loss function without explicit calculation of gradients. The correspondence holds exactly only in the limit of vanishing mutation scale, but we use numerics to show in \s{numerics} that it can be observed for neuroevolution done with finite mutations and gradient descent enacted with a finite timestep. We conclude in \s{conclusions}.

\section{Summary of main results}
\label{summary}

In this section we summarize the main analytic results of this paper. {These results are derived in \s{derivation}}.

Consider a neural network with $N$ parameters (weights and biases) $\x=\{x_1,\dots,x_N\}$, and a loss $\u$ that is a deterministic function of the network parameters. {The form of the network does not enter the proof, and so the result applies to neural networks of any architecture (we shall illustrate this point numerically by considering both deep and shallow nets).} The loss function may also depend upon other parameters, such as a set of training data, as in supervised learning, or a set of {actions} and states, as in reinforcement learning; the correspondence we shall describe applies regardless.

\subsection{Gradient descent}

Under the simplest form of gradient descent, the parameters $x_i$ of the network evolve according to numerical integration of
\beq
\label{gd}
\frac{\d x_i}{\d t} = -\alpha \frac{\partial \u}{\partial x_i}.
\eeq
Here time $t$ measures the progress of training, and $\alpha$ is the learning rate\c{rumelhart1995backpropagation,rumelhart1986learning,hecht1992theory,lecun1989backpropagation,chauvin1995backpropagation}.
 
\subsection{Neuroevolution}
\label{nev}

Now consider training the network by neuroevolution, defined by the following Monte Carlo protocol. 

\begin{enumerate}
\item Initialize the neural-network parameters $\x$ and calculate the loss function $\u$. Set time $t=0$.
\item Propose a change (or ``mutation'') of each neural-network parameter by an independent Gaussian random number of zero mean and variance $\sigma^2$, so that 
\beq
\x \to \x + \e,
\eeq
where $\e = \{\epsilon_1,\dots,\epsilon_N\}$ and $\epsilon_i \sim {\cal N}(0,\sigma^2)$.
\item Accept the mutation with the Metropolis probability $\min\left(1,\ee^{-\beta [\up-\u]}\right)$, and otherwise reject it. In the latter case we return to the original neural network. The parameter $\beta$ can be considered to be a reciprocal evolutionary temperature.
\item Increment time $t \to t+1$, and return to step 2.
\end{enumerate}

For finite $\beta$, and in the limit of small mutation scale $\sigma$, the parameters of the neural network evolve under this procedure according to the Langevin equation
\beq
\label{lang}
\frac{\d x_i}{\d t} = -\frac{\beta \sigma^2}{2} \frac{\partial \u}{\partial x_i} + \xi_i(t),
\eeq
where $\xi$ is a Gaussian white noise with zero mean and variance $\sigma^2$:
\beq
\av{\xi_i(t)}=0, \quad \av{\xi_i(t)\xi_j(t')}=\sigma^2 \delta_{ij} \delta(t-t').
\eeq
\eqq{lang} describes an evolution of the neural-network parameters $x_i$ that is equivalent to gradient descent with learning rate $\alpha = \beta \sigma^2/2$ in the presence of Gaussian white noise. Averaging over independent stochastic trajectories of the learning process (starting from identical initial conditions) gives
\bea
\label{lang_av}\frac{\d \av{x_i}}{\d t} = -\frac{\beta \sigma^2}{2} \frac{\partial \u}{\partial x_i},
\eea
which has the same form as the gradient descent equation \eq{gd}. Thus, when averaged over many independent realizations of the learning process, the neuroevolution procedure 1--4, with finite $\beta$, is equivalent in the limit of small mutation scale to gradient descent on the loss function.

In the case $\beta = \infty$, where mutations are only accepted if the loss function does not increase, the parameters of the network evolve according to the Langevin equation
 \bea
\label{lang2}
\frac{\d x_i}{\d t} = -\frac{\sigma}{\sqrt{2 \pi}}\frac{1}{|\nabla U(\x)|}\frac{\partial \u}{\partial x_i}+ \eta_i(t),
\eea
where $\eta$ is a Gaussian white noise with zero mean and variance $\sigma^2/2$:
\beq
\av{\eta_i(t)}=0, \quad \av{\eta_i(t)\eta_j(t')}=\frac{\sigma^2}{2} \delta_{ij} \delta(t-t').
\eeq
The form \eq{lang2} is different to \eq{lang}, because the gradient in the first term is normalized by the factor $|\nabla \u| = \sqrt{\sum_{i=1}^N (\partial \u/\partial x_i)^2}$, which serves as an effective coordinate-dependent rescaling {(or vector normalization)} of the timestep, but \eq{lang2} nonetheless describes a form of gradient descent on the loss function $\u$. Note that the drift term in \eq{lang2} is of lower order in $\sigma$ than the diffusion term (which is not the case for finite $\beta$). In the limit of small $\sigma$, \eq{lang2} describes an effective process in which uphill moves in loss cannot be made, consistent with the stochastic process from which it is derived.  

Averaged over independent realizations of the learning process, \eq{lang2} reads
\beq
\label{lang2_av}\frac{\d \av{x_i}}{\d t} = -\frac{\sigma}{\sqrt{2 \pi}}\frac{1}{|\nabla U(\x)|}\frac{\partial \u}{\partial x_i}.
\eeq
The results \eq{lang} and \eq{lang2} show that training a network by making random mutations of its parameters is, in the limit of small mutations, equivalent to noisy gradient descent on the loss function. 

Writing $\dot{U}(\x) = \dot{\x} \cdot \nabla \u$, using \eq{lang} and \eq{lang2}, and averaging over noise shows the evolution of the mean loss function under neuroevolution to obey, in the limit of small $\sigma$,
\bea
\label{mean_loss}
 \av{ \dot{U}(\x)}=
 \begin{cases}
    -\frac{\beta \sigma^2}{2} (\nabla \u)^2 & {\rm if} \, \beta \neq \infty\\
    -\frac{\sigma}{\sqrt{2 \pi}}|\nabla \u| & {\rm if} \, \beta = \infty
\end{cases},
\eea
equivalent to evolution under the noise-free forms of gradient descent \eq{lang_av} and \eq{lang2_av}.

 In the following section we derive the correspondence described here; in \s{numerics} we show that it can be observed numerically for non-vanishing mutations and finite integration steps.

\section{Derivation of the neuroevolution-gradient descent correspondence}
\label{derivation}

{We start by considering the quantity $\p$, the probability that a neural network has the set of parameters $\x$ at time $t$ under a given stochastic protocol. The time evolution of this quantity is governed by the {\em master equation}\c{risken1996fokker,van1992stochastic}, which in generic form reads
\bea
\label{me0}
\partial_t \p &=& \int \d\x' \left[ P(\x',t) W_{\x-\x'}(\x') \right.\nonumber \\&-&\left. \p W_{\x'-\x}(\x) \right].
\eea
The two terms in \eq{me} describe, respectively, gain and loss of the probability $\p$ (note that the probability to have {\em any} set of parameters is conserved, i.e. $\int \d \x\, \p=1$). The symbol $W_{\x'-\x}(\x)$ (sometimes written $\wp$) quantifies the probability of moving from the set of parameters $\x$ to the set of parameters $\x+ (\x' -\x)=\x'$, and encodes the details of the stochastic protocol. For the neuroevolution procedure defined in \s{nev} we write \eq{me0} as}
\bea
\label{me}
\partial_t \p &=& \int \d\e \left[ \pm \wm \right.\nonumber \\&-&\left. \p \w \right].
\eea
Here $\e$ denotes the set of random numbers (the ``mutation'') by which the neural-network parameters are changed; the integral $\int \d\e = \int_{-\infty}^\infty  \d \epsilon_i \cdots \int_{-\infty}^\infty  \d \epsilon_N$ runs over all possible choices of mutations; and 
\beq
\label{rate}
\w = p(\e) \min\left(1,\ee^{-\beta [\up-\u]}\right)
\eeq
is the rate for going from the set of parameters $\x$ to the set of parameters $\x+\e$. \eqq{rate} contains two factors. The first,
\beq
\label{probe}
p(\e) = \prod_{i=1}^N p(\epsilon_i) \quad {\rm with} \quad  p(\epsilon_i) = \frac{1}{\sqrt{2 \pi \sigma^2}}\ee^{-\frac{\epsilon_i^2}{2 \sigma^2}},
\eeq
quantifies the probability of proposing a set of Gaussian random numbers $\e$. The second factor, the Metropolis ``min'' function in \eq{rate}, quantifies the probability of accepting the proposed move from $\x$ to $\x + \e$; recall that $\u$ is the loss function.

We can pass from the master equation \eq{me} to a Fokker-Planck equation by assuming a small mutation scale $\sigma$, and expanding the terms in \eq{me} to second order in $\sigma$\c{risken1996fokker,van1992stochastic}. Thus
\beq
\label{expand1}
\pm \approx \left(1 - \en  +\frac{1}{2} (\en)^2 \right)\p
\eeq
and
\beq
\label{expand2}
\wm \approx \left(1 - \en  +\frac{1}{2} (\en)^2 \right)\w,
\eeq
where $\en =  \sum_{i=1}^N \epsilon_i \partial_i$ (note that $\partial_i \equiv \frac{\partial}{\partial x_i}$). Collecting terms resulting from the expansion gives
\bea
\label{fp1}
\partial_t \p &\approx& -\int \d\e  (\en) \p \w \nonumber \\
&+& \frac{1}{2}\int \d\e  (\en)^2 \p \w.
\eea
Taking the integrals inside the sums, \eq{fp1} reads
\bea
\label{fp2}
 \partial_t \p &\approx&-\sum_{i=1}^N \frac{\partial}{\partial x_i} \left( A_i(\x) \p\right) \nonumber\\
&+&\frac{1}{2}\sum_{i,j=1}^N \frac{\partial^2}{\partial x_i \partial x_j} \left( B_{ij}(\x) \p\right),
\eea
where
\beq
\label{ay}
A_i(\x)\equiv \int \d\e \, \epsilon_i \w,
\eeq
and
\beq
\label{bee}
B_{ij}(\x) \equiv \int \d\e \, \epsilon_i \epsilon_j \w.
\eeq

What remains is to calculate \eq{ay} and \eq{bee}, which we do in different ways depending on the value of the evolutionary reciprocal temperature $\beta$. 

\subsection{Finite $\beta$}
\label{finite}

In this section we consider finite $\beta$, in which case we can evaluate \eq{ay} and \eq{bee} using the results of Refs.\c{kikuchi1991metropolis,kikuchi1992metropolis} (making small changes in order to account for differences in proposal rates between those papers and ours). 

\eqq{ay} can be evaluated as follows ({writing $\u=U$ for brevity}):
\bea
\label{1}A_i(\x) &=& \int \d\e \, p(\e) \epsilon_i  \min\left(1,\ee^{-\beta [\up-\u]}\right) \\
\label{2}&\approx& \int \d\e \, p(\e) \epsilon_i  \min\left(1,1-\beta \en \uu\right) \\
\label{3}  &=& \int \d\e \, p(\e) \epsilon_i  \Theta(-\en \uu)\nonumber \\&+&\int \d\e \, p(\e) \epsilon_i  (1-\beta \en \uu) \Theta(\en \uu) \\
 \label{4}          &=& \int \d\e \, p(\e) \epsilon_i  \nonumber \\&-&\beta \int \d\e \, p(\e) \Theta(\en \uu) \sum_j  \epsilon_i\epsilon_j \partial_j \uu \\
 \label{5}          &=& -\beta \int \d\e \, p(\e) \Theta(\en \uu) \sum_j  \epsilon_i\epsilon_j \partial_j \uu \\
 \label{6}      &=& -\frac{\beta}{2} \int \d\e \, p(\e)  \sum_j  \epsilon_i\epsilon_j \partial_j \uu \\
  \label{7}     &=& -\frac{\beta \sigma^2}{2} \partial_i \uu.
       \eea
In these expressions $\Theta(x) =1$ if $x\geq 0$ and is zero otherwise. In going from \eq{1} to \eq{2} we have assumed that $\beta \en \u$ is small. This condition cannot be met for $\beta = \infty$; that case is treated in \s{inf}. In going from \eq{3} to \eq{4} we have used the result $\Theta(x) + \Theta(-x)=1$; the first integral in \eq{4} then vanishes by symmetry. {The second integral, shown in \eq{5}, can be turned into \eq{6} using the symmetry arguments given in~\cc{kikuchi1992metropolis}, which we motivate as follows.} Upon a change of sign of the integration variables, $\e \to -\e$, the value of the integral in \eq{5} is unchanged and it is brought to a form that is identical except for a change of sign of the argument of the $\Theta$ function. Adding the two forms of the integral removes the $\Theta$ functions, giving the form shown in \eq{6}, and dividing by 2 restores the value of the original integral. \eq{6} can be evaluated using standard results of Gaussian integrals.

\eqq{bee} can be evaluated in a similar way:
\bea
\label{b1}B_{ij}(\x)&=& \int \d\e \, p(\e) \epsilon_i \epsilon_j  \min\left(1,\ee^{-\beta [\up-\u]}\right) \\
\label{b2}&\approx& \int \d\e \, p(\e) \epsilon_i \epsilon_j   \min\left(1,1-\beta \en \uu\right) \\
\label{b3}  &=& \int \d\e \, p(\e) \epsilon_i \epsilon_j \Theta(-\en \uu)\nonumber \\&+&\int \d\e \, p(\e) \epsilon_i \epsilon_j  (1-\beta \en \uu)\Theta(\en \uu) \\
\label{b4}&\approx& \int \d\e \, p(\e) \epsilon_i \epsilon_j   \\
\label{b5}&=& \sigma^2 \delta_{ij}.
\eea
The $\approx$ sign in \eq{b4} indicates that we have omitted terms of order $\sigma^3$.

Inserting \eq{7} and \eq{b5} into \eq{fp2} gives us, to second order in $\sigma$, the Fokker-Planck equation 
\bea
\frac{\partial \p}{\partial t} &\approx& -\sum_{i=1}^N \frac{\partial}{\partial x_i} \left( -\frac{\beta \sigma^2}{2} \frac{\partial \u}{\partial x_i}  \p\right) \nonumber\\
&+&\frac{1}{2}\sum_{i=1}^N \frac{\partial^2}{\partial x_i^2 } \left( \sigma^2 \p\right).
\eea
This equation is equivalent~\footnote{The diffusion term is independent of $\x$, and so the choice of stochastic calculus is unimportant.} to the $N$ Langevin equations\c{risken1996fokker,van1992stochastic}
\bea
\label{langmain}
\frac{\d x_i}{\d t} = -\frac{\beta \sigma^2}{2} \frac{\partial \u}{\partial x_i} + \xi_i(t) \quad \forall i,
\eea
where $\xi$ is a Gaussian white noise with zero mean and variance $\sigma^2$:
\beq
\label{noisemain}
\av{\xi_i(t)}=0, \quad \av{\xi_i(t)\xi_j(t')}=\sigma^2 \delta_{ij} \delta(t-t').
\eeq
\eqq{langmain} describes an evolution of the neural-network parameters $x_i$ that is equivalent to gradient descent with learning rate $\alpha = \beta \sigma^2/2$, plus Gaussian white noise. Averaging over independent stochastic trajectories of the learning process (starting from identical initial conditions) gives
\bea
\frac{\d \av{x_i}}{\d t} = -\frac{\beta \sigma^2}{2} \frac{\partial \u}{\partial x_i},
\eea
which is equivalent to simple gradient descent on the loss function. 

\subsection{The case $\beta=\infty$}
\label{inf}

When $\beta = \infty$ we only accept mutations that do not increase the loss function. To treat this case we return to \eq{ay} and take the limit $\beta \to \infty$:
\beq
\label{ai}
A_i(\x) = \int \d \e \, p(\e) \epsilon_i  \Theta(-\en \u).
\eeq
We can make progress by introducing the integral form of the $\Theta$ function (see e.g.\c{gauss}),
\bea
\label{theta}
 \Theta(-\en \u)&=& \int_{-\infty}^{0} \d z\, \delta(z -\en \u) \\
 &=& \int_{-\infty}^0 \frac{\d z}{2 \pi} \int_{-\infty}^{\infty} \d \omega \, \ee^{{\rm i} \omega (z -\en \u)}. \nonumber
\eea
Then \eq{ai} reads
\bea
\label{int1}
\label{c1} A_i(\x) = \int_{-\infty}^0 \frac{\d z}{2 \pi} \int_{-\infty}^{\infty} \d \omega \, \ee^{{\rm i} \omega z }\, G_i^{(1)}\prod_{j\neq i}G_j^{(0)},
\eea
where the symbols
\bea
\label{gauss}
G_j^{(n)} \equiv \frac{1}{\sqrt{2 \pi \sigma^2}}\int_{-\infty}^\infty \d \epsilon_j\, \epsilon_j^n \, \ee^{-\epsilon_j^2/(2 \sigma^2) - {\rm i} \omega \epsilon_j \partial_j U}
\eea
are standard Gaussian integrals. Upon evaluating them as
\beq
G_j^{(0)}=\ee^{-\frac{1}{2}\sigma^2 \omega^2 (\partial_j U)^2}
\eeq
and 
\beq
G_i^{(1)}=-{\rm i} \omega \sigma^2 (\partial_i U) \ee^{-\frac{1}{2}\sigma^2 \omega^2 (\partial_i U)^2},
\eeq
\eq{int1} reads
\bea
\label{c2} A_i(\x) &=&-\int_{-\infty}^0 \frac{\d z}{2 \pi} \int_{-\infty}^{\infty} \d \omega \, {\rm i} \sigma^2 \omega (\partial_i U) \ee^{-\frac{1}{2}\omega^2 \sigma^2 |\nabla U|^2+{\rm i} \omega z } \nonumber \\
\label{c3} &=&\int_{-\infty}^0 \frac{\d z}{2 \pi}  \sqrt{2 \pi} \frac{ \partial_i U}{\sigma |\nabla U|^3} z \ee^{-z^2/(2 \sigma^2 |\nabla U|^2)}\\
\label{c4}&=&-\frac{\sigma}{\sqrt{2 \pi}}\frac{\partial_i U(\x)}{|\nabla U(\x)|}.
\eea
 The form \eq{c4} is similar to \eq{7} in that it involves the derivative of the loss function $\u$ with respect to $x_i$, but contains an additional normalization term, $|\nabla U|$. {This term is sometimes introduced as a form of regularization in gradient-based methods\c{pascanu2013difficulty}. Here, the form emerges naturally from the acceptance criterion that sees any move accepted if the move does not increase the loss function: as a result, the length of the step taken does not depend strongly on the size of the gradient.}
 
In the limit $\beta \to \infty$, \eq{bee} reads
\bea
B_{ij}(\x) &=& \int \d\e \, p(\e)\epsilon_i \epsilon_j  \Theta(-\en \u) \\
 &=& \frac{1}{2} \int \d\e \, p(\e)  \epsilon_i \epsilon_j \\
 \label{d1} &=&\frac{1}{2}\sigma^2 \delta_{ij},
\eea
upon applying the symmetry arguments used to evaluate \eq{5}. \eqq{d1} is half the value of the corresponding term for the case $\beta \neq \infty$, \eqq{b5}, because one term corresponds to Brownian motion in unrestricted space, the other to Brownian motion on a half-space.

Inserting \eq{c4} and \eq{d1} into \eq{fp2} gives a Fokker-Planck equation equivalent to the $N$ Langevin equations
 \bea
\label{lang2main}
\frac{\d x_i}{\d t} = -\frac{\sigma}{\sqrt{2 \pi}}\frac{1}{|\nabla U(\x)|}\frac{\partial \u}{\partial x_i}+ \eta_i(t) \quad \forall i,
\eea
where $\eta$ is a Gaussian white noise with zero mean and variance $\sigma^2/2$:
\beq
\label{noise2main}
\av{\eta_i(t)}=0, \quad \av{\eta_i(t)\eta_j(t')}=\frac{1}{2} \sigma^2\delta_{ij} \delta(t-t').
\eeq

{As an aside, we briefly consider the case of non-isotropic mutations, for which the Gaussian random update applied to parameter $i$ has its own variance $\sigma_i^2$, i.e. $\epsilon_i \sim {\cal N}(0,\sigma_i^2)$ in step 2 of the procedure described in \s{nev}. In this case the derivation above is modified to have $\sigma$ replaced by $\sigma_i$ in \eq{probe}. In the case of finite $\beta$ the equations \eq{langmain} and \eq{noisemain} retain their form with the replacement $\sigma \to \sigma_i$. In the case of infinite $\beta$, \eq{noise2main} retains its form with the replacement $\sigma \to \sigma_i$ and \eq{lang2main} reads
\beq
\frac{\d x_i}{\d t} = -\frac{\sigma_i}{\sqrt{2 \pi}}\frac{1}{|\tilde{\nabla} U(\x)|}\tilde{\partial}_i \u+ \eta_i(t),
\eeq
with $\tilde{\partial}_i \equiv \sigma_i \frac{\partial}{\partial x_i}$ and $|\tilde{\nabla} U| \equiv \sqrt{\sum_{j=1}^N \left(\tilde{\partial}_j U \right)^2}$. Non-isotropic mutations are used in covariance matrix adaptation strategies\c{hansen2006cma}. Those schemes also evolve the step size parameter dynamically, and to model this more general case one must make $\sigma$ a dynamical variable of the master equation, with update rules appropriate to the algorithm of interest.}

\section{Numerical illustration of the neuroevolution-gradient descent correspondence}
\label{numerics}

\begin{figure*}[] 
   \centering
   \includegraphics[width=\linewidth]{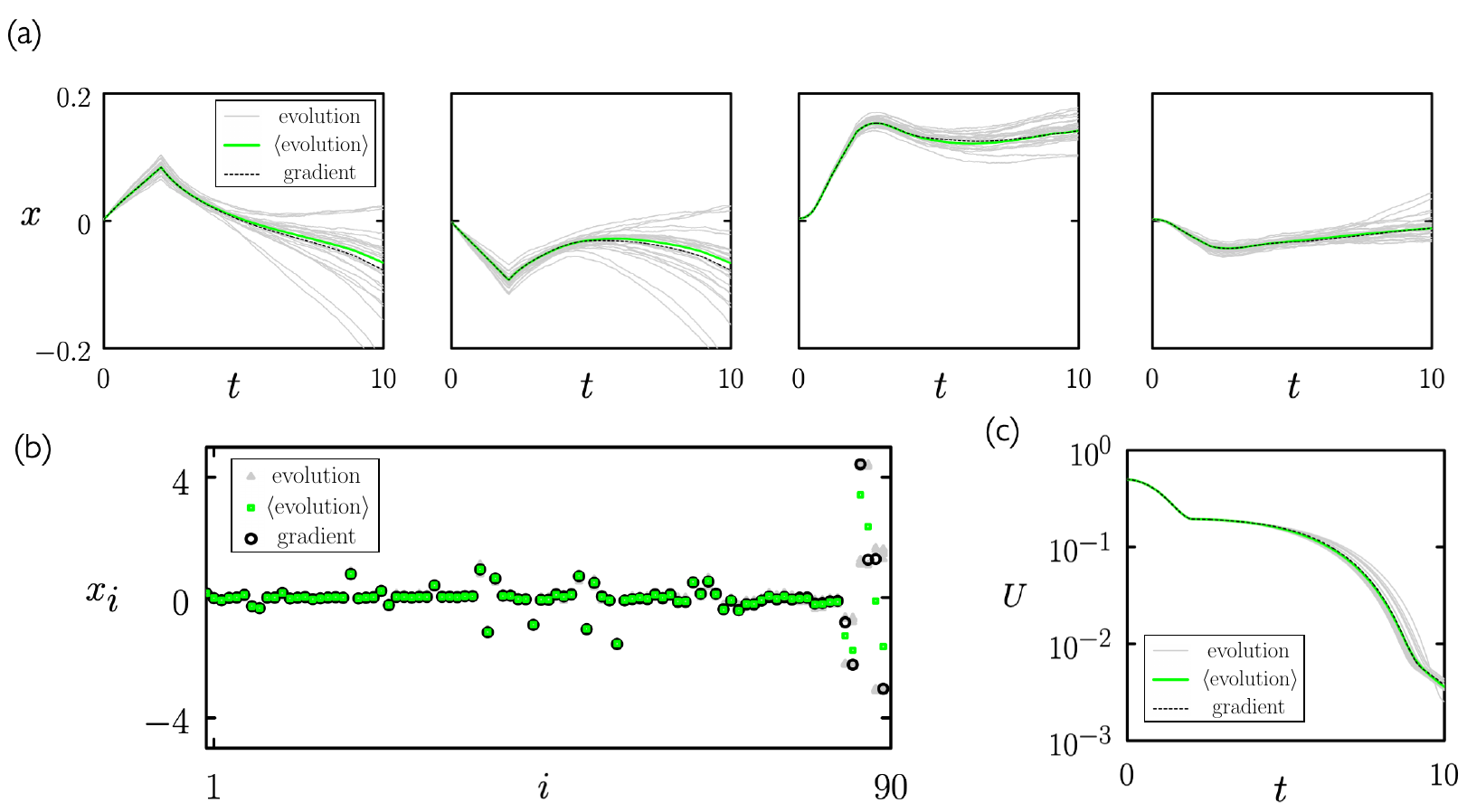} 
   \caption{Numerical illustration of the equivalence of gradient descent and neuroevolution in the case $\beta = \infty$. (a) Evolution with time of 4 of the 90 weights of the neural network \eq{net1} under the gradient descent equation \eqq{gd1} (black:``gradient'' stands for``gradient descent''), and under neuroevolution with mutation scale $\lambda=1/10$ [see \eq{set}]. Here and in subsequent panels we show 25 independent neuroevolution trajectories (grey, marked ``evolution'') and the average of 1000 independent trajectories (green, marked ``$\av{{\rm evolution}}$''). (b) All weights at time $t=10$ under the two methods. (c) Loss as a function of time.}
   \label{fig1}
\end{figure*}

\begin{figure*}[] 
   \centering
   \includegraphics[width=\linewidth]{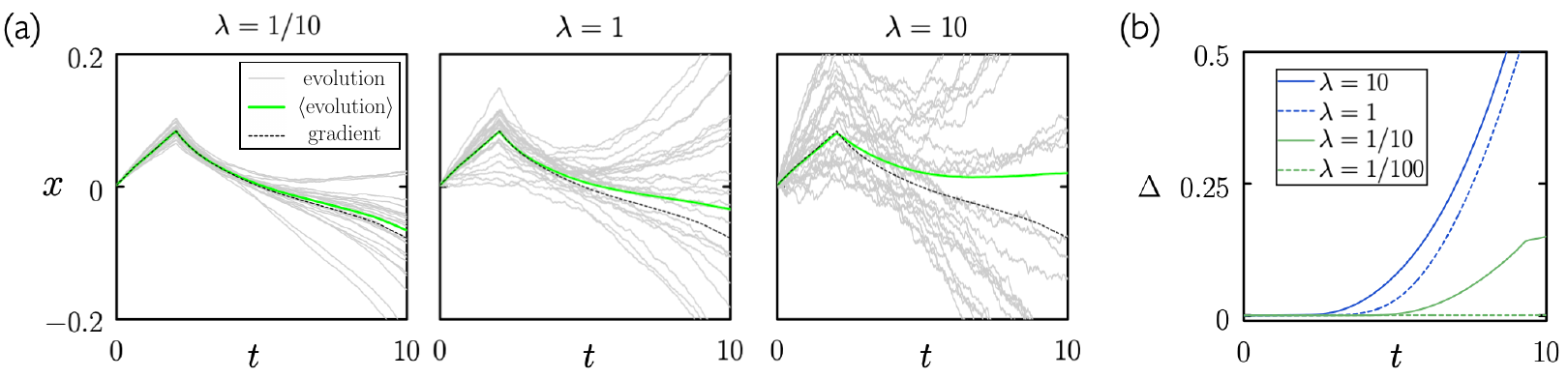} 
   \caption{The smaller the neuroevolution mutation scale, the closer the neuroevolution-gradient descent correspondence. (a) Time evolution of a single neural-network weight, for three different neuroevolution mutation scales [see \eq{set}]. (b) The quantity $\Delta(t)$, \eqq{delta_av}, is a measure of the { difference between} networks evolved under gradient descent and neuroevolution.}
   \label{fig2}
\end{figure*}

{In this section we demonstrate the neuroevolution-gradient descent correspondence numerically. We consider single-layer neural networks for the cases of infinite and finite $\beta$, and a deep network for the case of infinite $\beta$.}

\subsection{Shallow net, $\beta  = \infty$}
\label{shallow_infinite}

{In order to observe correspondence numerically, the neuroevolution mutation scale $\sigma$ must be} small enough that correction terms neglected in the expansion leading to \eq{fp2} and \eq{lang2main} are small. The required range of $\sigma$ is difficult to know in advance, but straightforward to determine empirically: {below the relevant value of $\sigma$, the results of neuroevolution will be statistically similar when scaled in the manner described below.}

We consider a simple supervised-learning problem in which we train a neural network to express the function $f_0(\theta) = \sin (2 \pi \theta)$ on the interval $\theta \in [0,1)$. We calculated the loss using $K=1000$ points on the interval,
\beq
U(\x) =\frac{1}{K} \sum_{j=0}^{K-1} \left[ f_{\x}(j/K)-f_0(j/K) \right]^2, 
\eeq
where 
\beq
\label{net1}
f_{\x}(\theta) = \sum_{i=0}^{M-1} x_{3 i+1} \tanh(x_{3 i +2} \theta+x_{3i+3})
\eeq
is the output of a single-layer neural network with one input node, one output node, $M=30$ hidden nodes, and $N=3M$ parameters $x_i$. These parameters are initially chosen to be Gaussian random numbers with zero mean and variance $\sigma_0^2=10^{-4}$. {The correspondence is insensitive to the choice of initial conditions, and we shall show that it holds for different choices of initial network.}

We performed gradient descent with learning rate $\alpha=10^{-5}$. {We chose the learning rate arbitrarily, and verified that the results of gradient-descent simulations were unchanged upon a changing learning rate by a factor of 10 and $1/10$.} We used Euler integration of the noise-free version of \eqq{lang2}, updating all weights $x_i$ at each timestep $t_{\rm gd}=1,2,\dots$ according to the prescription
\beq
\label{gd1}
x_i(t_{\rm gd}+1) = x_i(t_{\rm gd}) -\frac{\alpha}{|\nabla U(\x)|}\frac{\partial \u}{\partial x_i},
\eeq
where
\beq
\frac{\partial \u}{\partial x_i} =  \frac{2}{K} \sum_{j=0}^{K-1}  \left[ f_{\x}(j/K)-f_0(j/K) \right] \frac{\partial f_{\x}(j/K)}{\partial x_i},
\eeq
and 
\bea
 \frac{\partial f_{\x}(\theta)}{\partial x_i} = 
 \begin{cases}
    \tanh(\theta x_{i+1} +x_{i+2}) & {\rm if} \, i \, {\rm mod} \,3 = 1\\
   \theta x_{i-1} \,{\rm sech}^2(\theta x_{i} +x_{i+1})&  {\rm if} \,i \, {\rm mod} \,3 = 2\\
    x_{i-2}\, {\rm sech}^2(\theta x_{i-1} +x_i)              & {\rm if} \, i \, {\rm mod} \,3 = 0. \nonumber
\end{cases}
\eea

We did neuroevolution following the Monte Carlo procedure described in \s{nev}, in the limit $\beta = \infty$, i.e. we accepted only moves that did not increase the loss function. We chose the mutation scale 
\beq
\label{set}
\sigma = \lambda \alpha \sqrt{2 \pi},
\eeq
where $\lambda$ is a parameter. According to \eq{lang2} and \eq{gd1}, this prescription sets the neuroevolution timescale $t_{\rm evol}$ to be a factor $\lambda$  times that of the gradient-descent timescale. Thus one neuroevolution step corresponds to $\lambda$ integration steps of the gradient descent procedure. In figures, we compare gradient descent with neuroevolution as a function of common (scaled) time $t= \alpha  t_{\rm gd}= \alpha \lambda t_{\rm evol}$.

\begin{figure*}[] 
   \centering
   \includegraphics[width=\linewidth]{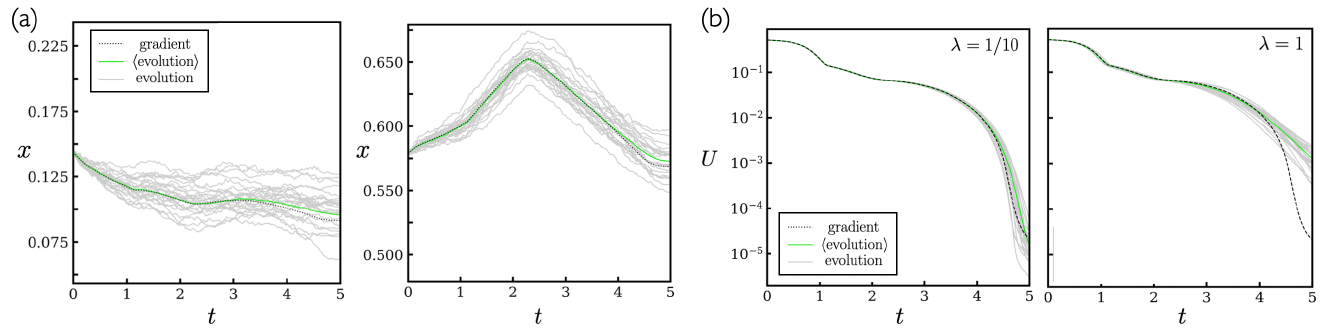} 
   \caption{{As \f{fig1}, now for a deep neural network with 8 hidden layers. Panel (a) shows 2 of the 7489 parameters of the network. Panel (b) shows the loss, for two different neuroevolution step-size parameters [see \eq{set2}].}}
   \label{fig_deep}
\end{figure*}

\begin{figure*}[] 
   \centering
   \includegraphics[width=\linewidth]{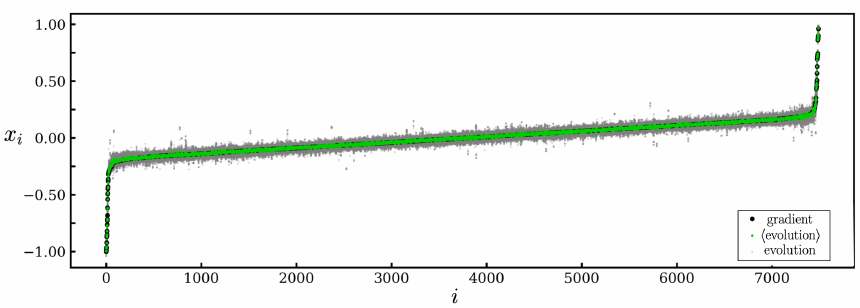} 
   \caption{{All 7489 parameters $x_i$ of the deep net of \f{fig_deep} at training time $t=5$. As predicted analytically, the neural network created by averaging (green symbols) over a noninteracting population of neuroevolutionary processes (gray symbols) is essentially the same network as that produced by gradient descent (black symbols). For clarity, weights are ordered by their final values under gradient descent.}}
   \label{fig_weights}
\end{figure*}

\begin{figure}[] 
   \centering
   \includegraphics[width=\linewidth]{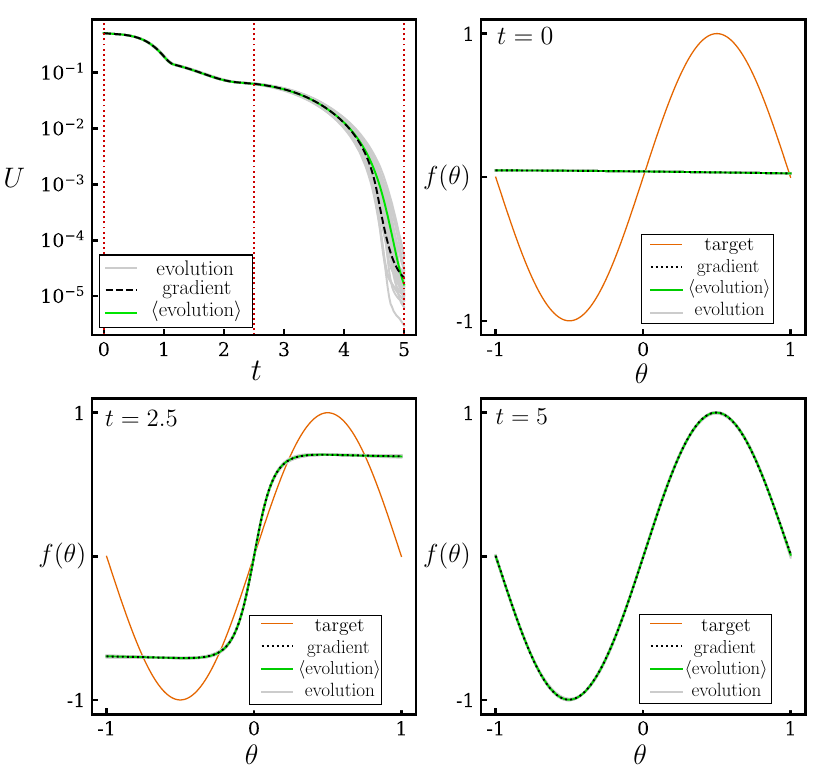} 
   \caption{{We illustrate the learning dynamics of \f{fig_deep} (for the case $\lambda=1/10$) and the scale of the loss function by showing a comparison between the target function $f_0(\theta) = \sin (\pi \theta)$ and the net function $f_{\x}(\theta)$ at three different training times.}}
   \label{fig_pred}
\end{figure}

In \f{fig1} (a) we show the evolution of {four} individual weights under neuroevolution {(using mutation scale $\lambda=1/10$)} and gradient descent (weights are distinguishable because they always have the same initial values). The correspondence predicted analytically can be seen numerically: individual neuroevolution trajectories (gray) fluctuate around the gradient descent result (black), and when averaged over individual trajectories the results of neuroevolution (green) approximate those of gradient descent. In \f{fig1}(b) we show the individual and averaged values of the weights of neuro-evolved networks at time $t=10$ compared to those of gradient descent. In general, the weights generated by averaging over neuroevolution trajectories approximate those of gradient descent, with some discrepancy seen in the values of the largest weights. In \f{fig1}(c) we show the loss under neuroevolution and gradient descent. As predicted by \eq{mean_loss}, averaged neuroevolution and gradient descent are equivalent. 

{In \f{fig_seed} we show similar quantities using a different choice of initial neural network; the correspondence between neuroevolution and gradient descent is again apparent.}

In \f{fig2}(a) we show the time evolution of a single weight of the network under gradient descent and neuroevolution, the latter for three sizes of mutation step $\sigma$. As $\sigma$ increases, the size of fluctuations of individual trajectories about the mean increase, as predicted by \eq{lang2}. As a result, more trajectories are required to estimate the average, and for fixed number of trajectories (as used here) the estimated average becomes less precise. In addition, as $\sigma$ increases, the assumptions underlying the correspondence derivation eventually break down, in which case the neuroevolution average will not converge to the gradient descent result even as more trajectories are sampled.

In \f{fig2}(b) we {show the mean-squared difference of the parameter vector of the model under gradient descent and neuroevolution,}
\beq
\label{delta_av}
\Delta(t) \equiv \frac{1}{N}\sum_{i=1}^N\left(x_i^{\rm gradient}(t)-\av{x_i^{\rm evolution}(t)} \right)^2.
\eeq
Here $N$ is the number of network parameters; $x_i^{\rm gradient}(t)$ is the time evolution of neural-network parameter $i$ under the gradient descent equation \eqq{gd1}; and $\av{x_i^{\rm evolution}(t)}$ is the mean value of neural-network parameter $i$ over the ensemble of neuroevolution trajectories. The smaller the neuroevolution step size, the smaller is $\Delta(t)$, and the closer the neuroevolution-gradient descent correspondence. 

In \f{fig3} we show the evolution with time of the loss for different mutation scales {(the left-hand plot is a reproduction of \f{fig1}(c))}. The trend shown is similar to that of the weights in \f{fig2}.

\subsection{{Deep net, $\beta  = \infty$}}
\label{deep_infinite}

One feature of the correspondence derivation is that the architecture of the neural network does not appear. As long as the loss $\u$ is a deterministic function of the neural-network parameters $\x$, correspondence between gradient descent and neuroevolution will be observed if the mutation scale is small enough. (It is likely that what constitutes ``small enough'' does depend on neural-network architecture, as well as the problem under study. {The required mutation scale can be determined empirically, even without access to gradient-descent results: when correspondence holds, the results of neuroevolution simulations will be statistically similar, when scaled as we have described}.) 

To demonstrate invariance to architecture we repeat the comparison of \s{shallow_infinite}, now using a deep neural network (we train the net to reproduce the target function $f_0(\theta) = \sin (\pi \theta)$ on the interval $\theta \in [-1,1]$). The network has 8 fully-connected hidden layers, each 32 nodes wide, and 7489 total parameters. As before we use tanh activations on the hidden nodes, and have one input node and one output.

Results are shown in \f{fig_deep}. In panel (a) we show the evolution of two parameters of the network, under gradient descent and for neuroevolution with step-size parameter $\lambda=1/10$ [see \eq{set2}]. As for the shallow net, the correspondence is apparent. Neuroevolution averages (green lines) are taken over 100 trajectories. In panel (b) we show the loss, for gradient descent and two different neuroevolution step-size parameters. As expected, the correspondence is more precise for smaller $\lambda$. {As before, for large enough $\lambda$ the correspondence breaks down: see \f{fig_break}.}

{In \f{fig_weights} we show all parameters of the deep net at training time $t=5$, under the two dynamics. We show the results of gradient descent in black, and independent neuroevolution trajectories in gray. As predicted analytically, the neuroevolution results fall either side of the gradient-descent result, and the network constructed by averaging over independent neuroevolution trajectories (green) is essentially identical to the network produced by gradient descent.}

{In \f{fig_pred} we illustrate the dynamics of learning and the scale of the loss function by showing a comparison between the target function $f_0(\theta) = \sin (2 \pi \theta)$ and the net function $f_{\x}(\theta)$. We show the latter at three different training times, for gradient descent and neuroevolution trajectories.}

{\subsection{Shallow net, finite $\beta$}}
\label{shallow_finite}
\begin{figure*}[] 
   \centering
   \includegraphics[width=\linewidth]{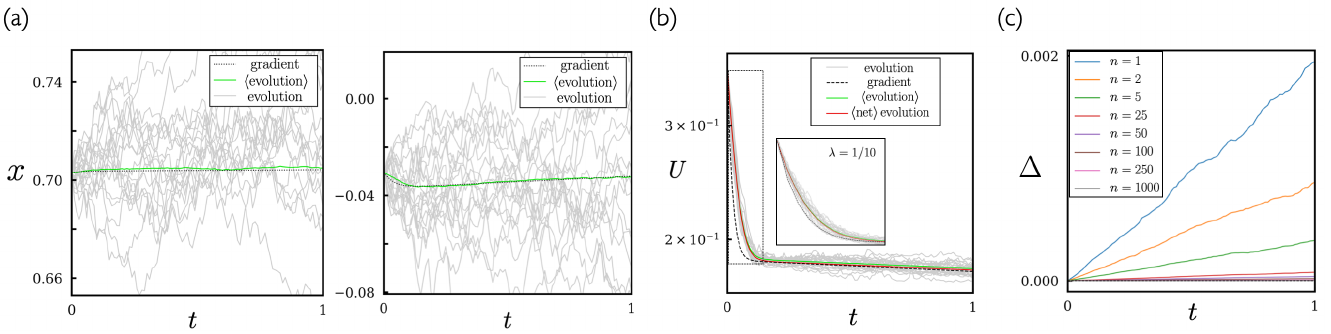} 
   \caption{{Illustration of the neuroevolution-gradient descent correspondence for the case of finite $\beta$. Panel (a) shows two parameters of the network, panel (b) shows the loss, and panel (c) shows the parameter $\Delta$, \eqq{delta_av}, a measure of the difference between the average network produced by neuroevolution and the network produced by gradient descent.} } 
   \label{fig_finite1}
\end{figure*}

In this section we illustrate the gradient descent-neuroevolution correspondence for finite $\beta$. We consider the same supervised-learning problem as in \s{shallow_infinite}, and set the network width to 256 nodes. We did neuroevolution with the Metropolis acceptance rate with reciprocal temperature parameter $\beta = 10^3$. {This choice is arbitrary, but is representative of a wide range of finite values of $\beta$.} Finite-temperature simulations are common in particle-based systems\c{frenkel2001understanding}. Here, temperature has no particular physical significance, but comparing simulations done at finite and infinite $\beta$ makes the point that different choices of neuroevolution acceptance rate result in a dynamics equivalent to different gradient-descent protocols. 

We did gradient descent using the integration scheme
\beq
\label{gd2}
x_i(t_{\rm gd}+1) = x_i(t_{\rm gd}) -\alpha\frac{\partial \u}{\partial x_i},
\eeq
where $\alpha = 10^{-4}$ is the learning rate. Comparing \eq{lang} and \eq{gd2}, we set the neuroevolution mutation scale to be
\beq
\label{set2}
\sigma = \lambda \sqrt{2 \alpha/ \beta},
\eeq
where $\lambda$ is a parameter. Thus one neuroevolution step corresponds to $\lambda$ integration steps of the gradient descent procedure. In figures, we compare gradient descent with neuroevolution as a function of common (scaled) time $t= \alpha  t_{\rm gd}= \alpha \lambda t_{\rm evol}$.

Results are shown in \f{fig_finite1}, for step-size parameter $\lambda=1$. In panel (a) we show the evolution with time of two of the weights of the network. The noise associated with neuroevolution at this value of $\beta$ is considerable: individual trajectories (grey lines) bear little resemble to the gradient-descent result (black line). However, the population average (green line, average of $n=1000$ trajectories) shows the expected correspondence. The correspondence is less precise than that shown in \s{shallow_infinite} because we use a larger effective step size and each trajectory is much noisier than its infinite-$\beta$ counterpart.

In \f{fig_finite1}(b) we show the loss, with line colors corresponding to the quantities of panel (a). In addition, we show the loss of the average network produced by neuroevolution (red line), $U(\av{\x})$, which, if correspondence holds, should be equal to $\av{U(\x)}$ (green line). The initial fast relaxation of the loss (the boxed region) shows a difference between gradient descent and averaged neuroevolution results; doing neuroevolution for smaller step-size parameter $\lambda=1/10$ (inset) reduces this difference, as expected. 

In panel (c) we show the parameter $\Delta$, \eqq{delta_av}, a measure of the difference between the average network $\av{x}$ produced by neuroevolution and the network produced by gradient descent, as a function of $n$, the number of trajectories included in the average. If correspondence holds, this quantity should vanish in the limit of large $n$; the observed trend is consistent with this behavior.

In \f{fig_finite2} we compare a gradient-descent trajectory with a set of neuroevolution trajectories, periodically resetting the latter to the gradient-descent solution. The periodic resetting tests the correspondence for a range of initial conditions. The correspondence between gradient descent and the averaged neuroevolution trajectory is approximate (averages were taken over 152 trajectories, fewer than in \f{fig_finite2}) but apparent.
~\\
\section{Conclusions}
\label{conclusions}

We have shown analytically that training a neural network by neuroevolution of its weights is equivalent, in the limit of small mutation scale, to noisy gradient descent on the loss function. Conditioning neuroevolution on the Metropolis acceptance criterion at finite evolutionary temperature is equivalent to a noisy version of simple gradient descent, while at infinite reciprocal evolutionary temperature the procedure is equivalent to clipped gradient descent on the loss function. Averaged over noise, the evolutionary procedures correspond to forms of gradient descent on the loss function. This correspondence is described by Equations \eq{lang}, \eq{lang_av}, \eq{lang2} and \eq{lang2_av}. 

{Correspondence in the sense described above means that each neural-network parameter evolves the same way as a function of time under the two dynamics. Correspondence implies that the convergence properties of the two methods are the same (see e.g. \f{fig_deep}(b)) and that the neural networks produced by the same methods are the same (see e.g. \f{fig_weights}). The generalization properties of those networks will then also be the same.}

The correspondence is formally exact only in the limit of zero mutation scale, {and holds approximately for small but finite mutations. It will fail when the assumptions underlying the derivation are violated, such as when the terms neglected in \eq{expand1} and \eq{expand2} are not small, or when the passage from \eq{1} to \eq{2} is not valid because the change $\beta \up-\beta \u$ is not small. {It is straightforward to determine empirically where correspondence holds, even without access to gradient-descent results: the results of neuroevolution, with time scaled as described, will be statistically similar when the mutation size is small enough.} The time duration for which the correspondence holds increases with decreasing mutation scale (see e.g. \f{fig2}(b)). We have shown here that the correspondence can be observed for a range of mutation scales, and for different neural-net architectures.}

{More generally,} several dynamical regimes are contained within the neurevolution master equation \eq{me}, according to the scale $\sigma$ of mutations~\footnote{Choosing $\sigma$ to be a parameter in a genetic search scheme would allow exploration of the dynamical diversity contained within \eqq{me}.}: for vanishing $\sigma$, neurevolution is formally noisy gradient descent on the loss function; for small but nonvanishing $\sigma$ it approximates noisy gradient descent enacted by explicit integration with a finite timestep; for larger $\sigma$ it enacts a dynamics different to gradient descent, but one that can still learn; and for sufficiently large $\sigma$ the steps taken are too large for learning to occur on accessible timescales. An indication of these various regimes can be seen in Figs.~\ref{fig2} and~\ref{fig3}. 

Separate from the question of its precise temporal evolution, the master equation \eq{me} has a well-defined stationary distribution $\rho_0(\x)$. Requiring the brackets on the right-hand of \eq{me} to vanish ensures that $P(\x,t) \to \rho_0(\x)$ becomes independent of time. Inserting \eq{rate} into \eq{me} and requiring normalization of $\rho_0(\x)$ reveals the stationary distribution to be the Boltzmann one, $\rho_0(\x) = \ee^{-\beta U(\x)}/\int \d \x' \ee^{-\beta U(\x')}$. For finite $\beta$ the neuroevolution procedure is ergodic, and this distribution will be sampled given  sufficiently long simulation time. For $\beta \to \infty$ we have $\rho_0(\x) \to \delta{\left( U(\x) - U_0 \right)}$, where $U_0$ is the global energy minimum; in this case the system is not ergodic (moves uphill in $U(\x)$ are not allowed) and there is no guarantee of reaching this minimum. 

We have focused on the simple limit of the set of neuroevolution algorithms, namely a non-interacting population of neural networks that experience sequential probabilistic mutations of their parameters. {We have illustrated the correspondence at the level of population averages, Equations \eq{lang_av} and \eq{lang2_av}. However, no communication between individuals is required, and each individually observes the correspondence defined by Equations \eq{lang} and \eq{lang2}.}

{Our results are also relevant to population-based genetic algorithms in which members of the population are periodically reset to the identities of the individuals with lowest loss values\c{GA,GA2,montana1989training}. For instance, when correspondence holds, individuals in the neuroevolution populations considered in this paper have an averaged loss equal to that of the corresponding gradient descent algorithm. Therefore, some individuals must have loss {\em less} than that of the corresponding gradient descent algorithm (see e.g. \f{fig1}(a), and \f{fig_deep}(b) for the case $\lambda=1/10$). This observation indicates the potential for such methods to be competitive with gradient-descent algorithms.}

The neuroevolution-gradient descent correspondence we have identified follows from that between the overdamped Langevin dynamics and Metropolis Monte Carlo dynamics of a particle in an external potential\c{kikuchi1991metropolis,kikuchi1992metropolis}. Our work therefore adds to the existing set of connections between machine learning and statistical mechanics\c{engel2001statistical,bahri2020statistical}, and continues a trend in machine learning of making use of old results: the stochastic and deterministic algorithms considered here come from the 1950s\c{metropolis1953equation,hastings1970monte} and {1970s}\c{linnainmaa1976taylor,werbos1982applications,rumelhart1995backpropagation,rumelhart1986learning,hecht1992theory,lecun1989backpropagation}, and are connected by ideas developed in the 1990s\c{kikuchi1991metropolis,kikuchi1992metropolis}.

{\em Acknowledgments} -- This work was performed as part of a user project at the Molecular Foundry, Lawrence Berkeley National Laboratory, supported by the Office of Science, Office of Basic Energy Sciences, of the U.S. Department of Energy under Contract No. DE-AC02--05CH11231. This work used resources of the National Energy Research Scientific Computing Center (NERSC), a U.S. Department of Energy Office of Science User Facility operated under Contract No. DE-AC02-05CH11231. I.T. acknowledges funding from the National Science and Engineering Council of Canada and carried out work at the National Research Council of Canada under the auspices of the AI4D Program.


%


\renewcommand{\theequation}{S\arabic{equation}}
\renewcommand{\thefigure}{S\arabic{figure}}
\renewcommand{\thesection}{S\arabic{section}}
\setcounter{figure}{0}

\begin{figure*}[] 
   \centering
   \includegraphics[width=\linewidth]{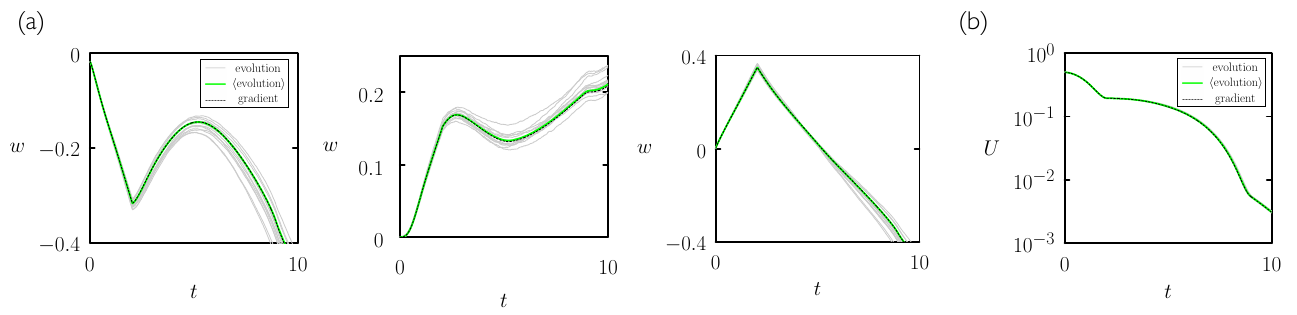} 
   \caption{{As \f{fig1}(a) and (c), but with simulations initiated from a different randomly-chosen neural network; the neuroevolution-gradient descent correspondence is again apparent. Here we show three weights (a) and the loss (b).}} 
   \label{fig_seed}
\end{figure*}

\begin{figure*}[] 
   \centering
   \includegraphics[width=\linewidth]{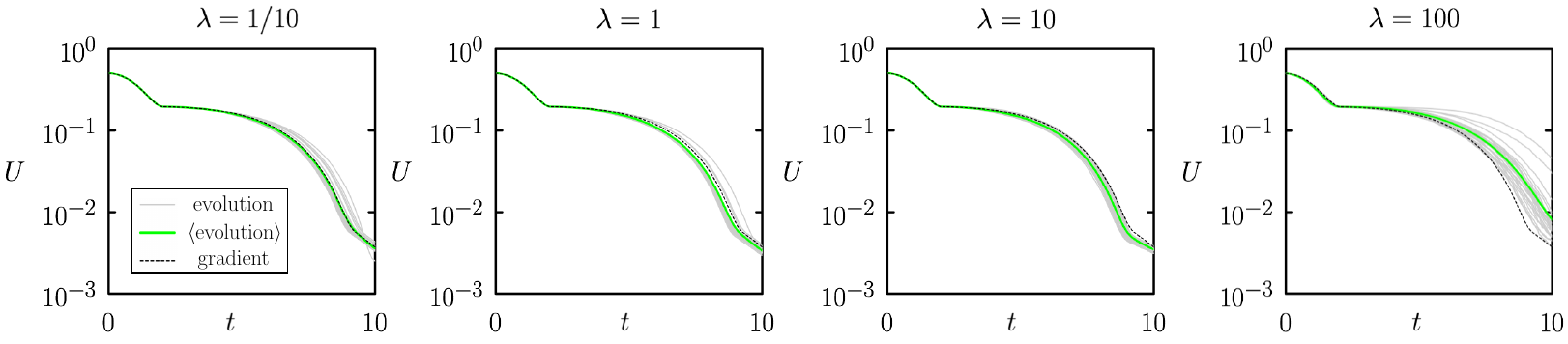} 
   \caption{As \f{fig2}(a), but showing the evolution of the loss function $U(\x)$ with time. For small mutations $\lambda$ [see \eqq{set}] the neuroevolution-gradient descent correspondence is apparent; for larger mutations the dynamics of neuroevolution and gradient descent are different, but neurevolution can still learn. For sufficiently large mutations, neuroevolution will cease to learn.} 
   \label{fig3}
\end{figure*}

\begin{figure*}[] 
   \centering
   \includegraphics[width=0.7\linewidth]{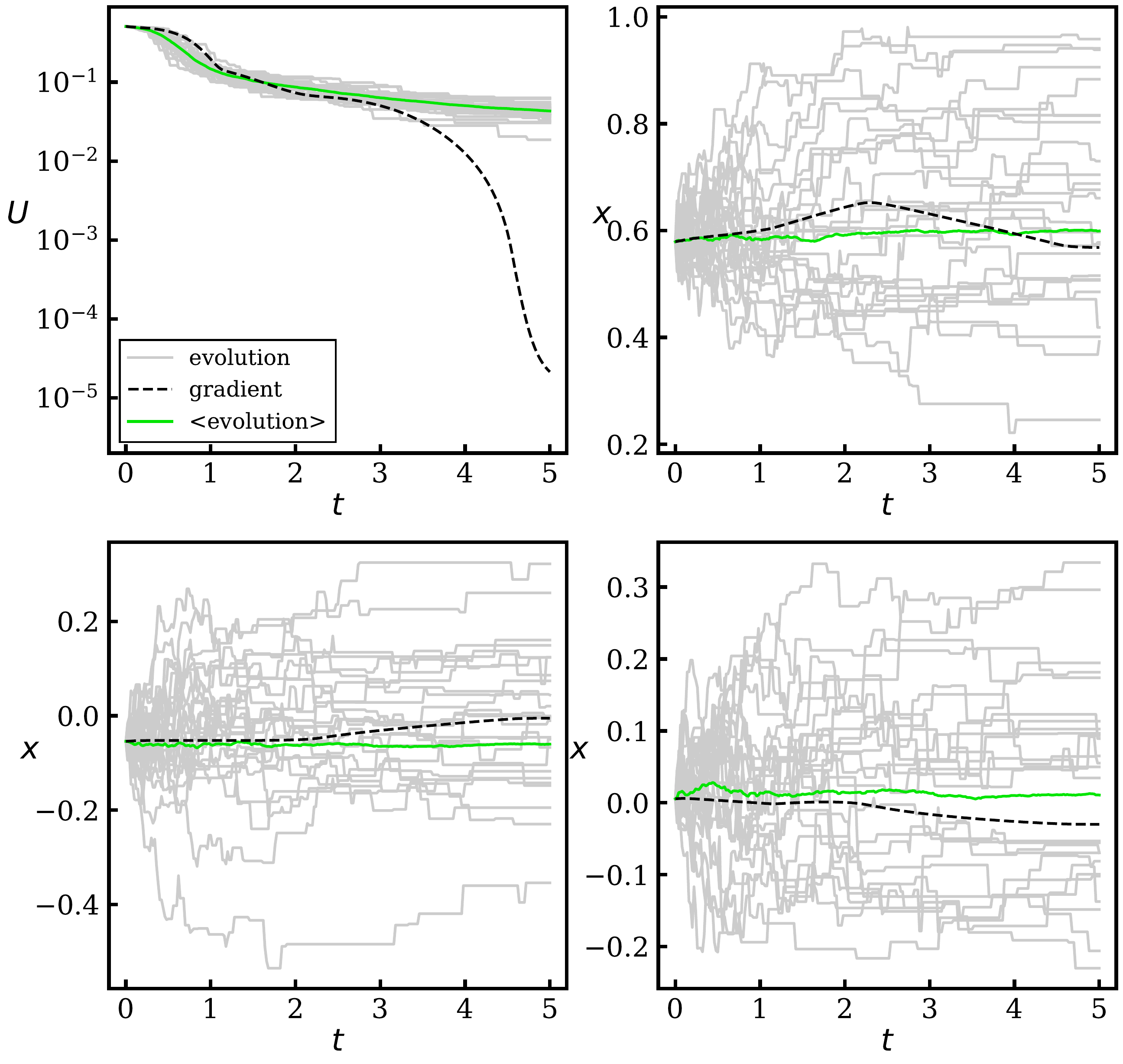} 
   \caption{{As \f{fig_deep}, but for mutations large enough ($\lambda=100$) that correspondence has broken down.}}
   \label{fig_break}
\end{figure*}

\begin{figure*}[] 
   \centering
   \includegraphics[width=0.8\linewidth]{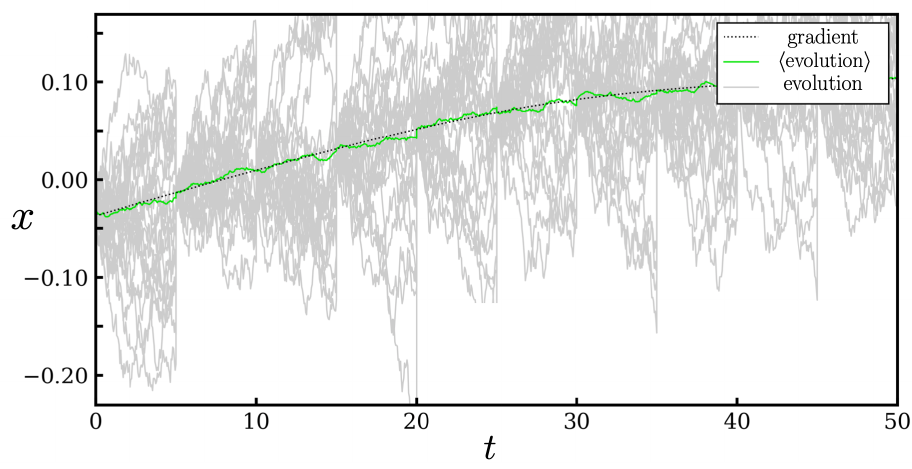} 
   \caption{{Supplement to \f{fig_finite1}: Illustration of the neuroevolution-gradient descent correspondence for the case of finite $\beta$, with periodic resetting (every 5 time units) of the neurevolution trajectories. The panel shows one weight of the network.} } 
   \label{fig_finite2}
\end{figure*}

\end{document}